%% file: main.tex
\documentclass[10pt,twocolumn,letterpaper]{article}

\usepackage[pagenumbers]{cvpr} 

\input{preamble}

\definecolor{cvprblue}{rgb}{0.21,0.49,0.74}
\usepackage[pagebackref,breaklinks,colorlinks,allcolors=cvprblue]{hyperref}

\def\confName{CVPR}
\def\confYear{2027}

\title{Training-Free Domain Generation for One-Shot Federated Learning with Server-Side Pre-trained Diffusion Models}

\author{
  Yixuan Tong$^{1}$ \quad
  Yuanguo Bi$^{1}$ \quad
  Mingjian Zhi$^{1}$ \quad
  Wenchao Xu$^{2}$ \quad
  Yuan Wu$^{3}$ \\
  $^{1}$Northeastern University, China \quad
  $^{2}$Hong Kong University of Science and Technology, China \\
  $^{3}$University of Macau, China
}

\begin{document}
\maketitle
\input{sec/0_abstract}
\input{sec/1_intro}
\input{sec/2_related}
\input{sec/3_method}
\input{sec/4_experiments}
\input{sec/5_conclusion}
\input{sec/6_LLM}
{
  \small
  \bibliographystyle{ieeenat_fullname}
  \bibliography{main}
}

\appendix
\input{sec/X_suppl}

\end{document}

%% file: preamble.tex
\usepackage{multirow}
\usepackage{amsthm}

\usepackage{algorithm}
\usepackage{algpseudocode}

\algtext*{EndFor}
\algtext*{EndWhile}
\algtext*{EndIf}

\newtheorem{theorem}{Theorem}

\usepackage{xcolor}

%% file: sec/0_abstract.tex
\begin{abstract}
  One-shot federated learning (OSFL) has attracted increasing attention for its ability to substantially reduce communication overhead by limiting client-server interaction to a single communication round.
  Nevertheless, this limited interaction makes learning from heterogeneous client distributions challenging.
  Diffusion-based OSFL approaches seek to address this challenge by synthesizing data that approximate client distributions.
  However, existing diffusion-based approaches often impose substantial computational costs on clients and lack effective visual guidance to generate data aligned with client distributions.
  To address these limitations, we propose \mbox{FedDoG}, a training-free \textbf{Do}main \textbf{G}eneration framework that leverages server-side pre-trained diffusion models to synthesize data aligned with client domains.
  FedDoG extracts compact domain representations using a frozen image encoder, substantially reducing client-side computational and communication costs.
  These representations are incorporated into the diffusion model through a pre-trained image-conditioning adapter, thereby providing explicit visual guidance for domain generation.
  To further improve the quality of synthetic data, FedDoG adapts the strength of visual conditioning according to domain concentration and filters generated samples based on class consistency and predictive entropy.
  Experiments on three benchmarks demonstrate that FedDoG generates higher-quality synthetic images and outperforms existing OSFL baselines in classification accuracy.
  Meanwhile, FedDoG reduces client upload and computation time by 99.9\% and 91.5\%, respectively.
  The code is available at \url{https://github.com/YXTong17/FedDoG}.
\end{abstract}

%% file: sec/1_intro.tex
\section{Introduction}
\label{sec:introduction}

The growing adoption of data-driven intelligence in healthcare, autonomous driving, and smart devices has raised increasing concerns about data privacy~\cite{shahid2022data,bendiab2023autonomous,aouedi2024survey}.
Federated Learning (FL)~\cite{mcmahan2017communication, kairouz2021advances} enables privacy-preserving collaborative model training across distributed clients without sharing raw data.
However, conventional FL requires repeated client-server communication, resulting in substantial communication overhead~\cite{amato2025towards}.
One-shot federated learning (OSFL)~\cite{jhunjhunwala2024fedfisher, zhang2022dense, guan2025capture} alleviates this burden by restricting client-server communication to a single round.

Despite its communication efficiency, OSFL still faces several challenges.
Some approaches aggregate independently trained local models~\cite{jhunjhunwala2024fedfisher,liu2024fedlpa}, but heterogeneous client distributions can lead to divergent local models, making effective aggregation challenging.
Others employ knowledge distillation~\cite{zhang2022dense,dai2024enhancing} to avoid direct parameter aggregation.
However, both approaches rely on sharing locally trained models, posing potential privacy risks.
Alternatively, some methods construct classifiers from compact feature statistics~\cite{guan2025capture, legate2023guiding}, but their reliance on frozen features limits task-specific representation learning.
Recently, diffusion-based OSFL methods~\cite{chen2025fedbip, yang2024exploring} have leveraged pre-trained diffusion models to synthesize client-conditioned data for server-side global model training.
These methods can construct a balanced synthetic training set across client domains, thereby mitigating biases arising from heterogeneous client distributions.

However, existing diffusion-based OSFL methods still face several critical limitations.
Firstly, some methods require local model training~\cite{yang2025oneshot, qiu2025fedsde}, while others deploy pre-trained diffusion models on clients and perform backpropagation through these models to optimize text conditioning representations~\cite{yang2024feddeo, chen2025fedbip}.
Even with the diffusion models kept frozen, such optimization imposes substantial computational and memory demands on resource-constrained clients.
Secondly, existing methods lack effective visual guidance for generating data aligned with heterogeneous client distributions.
Some approaches rely on learnable text conditioning representations to guide diffusion models~\cite{yang2024feddeo, yang2024exploring}, but these representations may not adequately convey client-specific visual characteristics.
Although other methods additionally use image-derived latents to initialize the diffusion process~\cite{chen2025fedbip}, such initialization do not provide explicit visual conditioning throughout denoising.
Thirdly, most diffusion-based OSFL methods rely on labeled client data, while others require labeled server data to construct class prototypes for client pseudo-labeling~\cite{yang2024exploring}.
Such dependence on labeled data limits their applicability when annotations are scarce or unavailable.
These limitations raise the following question:

\emph{How can pre-trained diffusion models generate synthetic data aligned with client distributions while keeping client computation and communication lightweight in OSFL?}

In this paper, we propose FedDoG, a training-free \textbf{Do}main \textbf{G}eneration framework for OSFL that leverages server-side pre-trained diffusion models to synthesize data aligned with heterogeneous client distributions.
Instead of relying on local model training or optimization, FedDoG extracts compact visual \emph{domain representations} using a frozen image encoder, substantially reducing client-side computation and communication overhead.
To overcome the limitation of text conditioning, we employ a pre-trained \mbox{IP-Adapter}~\cite{ye2023ip} to incorporate visual domain representations as explicit guidance during denoising.
Such visual conditioning complements text-based class semantics with client-specific visual characteristics without requiring optimization of the diffusion model or its conditioning modules.
Since domain representations are extracted without class labels, FedDoG requires no labeled data from either clients or the server, substantially broadening its applicability.
When class labels are available, FedDoG can further leverage class-level domain representations to provide finer-grained visual guidance.
To further improve synthetic data quality, we design \emph{domain-aware adaptive conditioning} to adjust visual conditioning strength based on domain concentration.
We also propose a \emph{synthetic data filtering} strategy based on class consistency and predictive entropy to select reliable samples for server-side global model training.
Our contributions can be summarized as follows:
\begin{itemize}
  \item We propose FedDoG, a training-free domain generation framework for OSFL that synthesizes domain-aligned data using frozen server-side diffusion models and a single lightweight upload from each client.
  \item We present a lightweight domain representation extraction mechanism to summarize client-specific visual characteristics and leverage the resulting representations to provide explicit visual guidance throughout denoising.
  \item We design domain-aware adaptive conditioning and synthetic data filtering strategies to enable robust generation of high-quality synthetic data under heterogeneous client distributions.
  \item We conduct experiments on three OSFL benchmarks and demonstrate superior performance against existing methods with substantially lower client communication and computation overhead.
\end{itemize}

%% file: sec/2_related.tex
\section{Related Work}
\label{sec:related_work}

\subsection{One-Shot Federated Learning}

OSFL offers a promising approach to communication-efficient distributed learning.
Existing approaches combine independently trained local models through parameter alignment or posterior aggregation~\cite{yurochkin2019bayesian, singh2020model, jhunjhunwala2024fedfisher, liu2024fedlpa}, or use model ensembles and distillation with public or synthetic data~\cite{guha2019one, li2020practical, zhang2022dense, dai2024enhancing}.
Others construct compact local datasets or train generators for global learning~\cite{zhou2020distilled, song2023federated, heinbaugh2023data}, retaining local training or data-construction costs.
Pre-trained models further reduce optimization through class prototypes~\cite{legate2023guiding}, analytic classifiers~\cite{fani2024accelerating, he2025afl}, and feature distribution modeling~\cite{guan2025capture, goswami2026covariances, beitollahi2025foundation}.
However, their single-round formulations primarily adapt classifiers within frozen feature spaces, limiting task-specific representation learning.
In contrast, FedDoG enables single-round learning from heterogeneous clients without local model training or dataset optimization, while supporting task-specific representation learning in the global model.

\subsection{Diffusion-based One-Shot Federated Learning}

Recently, pre-trained diffusion models have been leveraged in OSFL to synthesize client-conditioned data for server-side global model training.
However, some methods deploy pre-trained diffusion models on clients to optimize generation conditions~\cite{yang2024feddeo, chen2025fedbip}, while others train diffusion models locally~\cite{qiu2025fedsde}, increasing computational and memory demands on resource-constrained clients.
FedLMG~\cite{yang2025oneshot} deploys the diffusion model on the server side, but still requires clients to perform local model training to support one-shot knowledge transfer.
FedDISC~\cite{yang2024exploring} avoids local training, but relies on labeled server data to construct class prototypes for client pseudo-labeling.
FGL~\cite{zhang2023federated} only collects local sample features, but textual descriptions may omit visual details relevant to client domains.
Its instance-level uploads also increase communication overhead and pose potential privacy risks.
FedDoG jointly addresses these limitations by using compact, aggregated visual representations to guide server-side generation, requiring only frozen feature extraction on clients.

%% file: sec/3_method.tex
\section{Preliminaries}
\label{sec:preliminaries}

\subsection{Diffusion Models}

Diffusion models generate images by iteratively denoising a noise signal through a sequence of refinement steps.
Stable Diffusion (SD)~\cite{rombach2022high} is a representative method that employs a U-Net architecture to iteratively refine latent variables.
The denoising process is conditioned on text features produced by a text encoder, and these features are incorporated into the U-Net via attention mechanisms.
By performing denoising in a compact latent space rather than pixel space, latent diffusion models enable efficient high-resolution image generation.

\subsection{IP-Adapter}

To enable image-conditioned generation in pre-trained diffusion models, IP-Adapter~\cite{ye2023ip} incorporates image features through cross-attention decoupled from text conditioning.
Image features extracted from a frozen CLIP image encoder are incorporated into the U-Net as image-conditioned key-value pairs.
The decoupled cross-attention is formulated as:
\begin{equation}
  Z = \mathrm{Attention}(Q, K, V) + \lambda \cdot \mathrm{Attention}(Q, K', V'),
  \label{eq:ip_adapter}
\end{equation}
where $Q$ denotes queries from the current U-Net latent features, while $(K,V)$ are key-value pairs computed from text features and $(K',V')$ are key-value pairs computed from image features.
The scalar $\lambda$ controls the strength of image/domain conditioning relative to text guidance.

\subsection{Problem Setup}

We consider an OSFL setting with a central server and $N$ clients, where the goal is to learn a classifier over $C$ classes.
Client $i$ holds a private dataset $\mathcal{D}_i$ of $n_i$ samples, and the local data distributions may vary across clients.
In this work, a \emph{domain} refers to a data sub-distribution characterized by distinctive visual or semantic properties.
By default, we adopt client-level granularity, treating each client's local data distribution as a domain.
Communication is limited to a single round, after which the server learns a global classifier from client-provided information without accessing the raw local data.


\section{Method}
\label{sec:method}

\subsection{Overview}
\label{sec:method_overview}

We propose FedDoG, a training-free server-side generation framework that leverages client domain representations to condition pre-trained diffusion models for synthesizing data that approximate client distributions.
FedDoG consists of four stages:
(1) domain representation extraction,
(2) domain-aware adaptive conditioning,
(3) server-side domain generation,
and (4) synthetic data filtering.
The complete one-shot protocol is summarized in \cref{alg:fedog}.
Each client performs only frozen feature extraction and uploads a compact domain representation and concentration score once.
The server then converts these statistics into adaptive conditioning signals, generates and filters domain-specific samples, and trains the global classifier entirely on the resulting synthetic dataset.

\begin{algorithm}[t]
  \caption{FedDoG for One-Shot Federated Learning}
  \label{alg:fedog}
  \begin{algorithmic}[1]
    \Require Client datasets $\{\mathcal{D}_i\}_{i=1}^{N}$, class prompts $\{t_c\}_{c=1}^{C}$, frozen CLIP encoder $E$, pre-trained diffusion model $G$, generation budget $B$, entropy threshold $\tau$
    \Ensure Global classifier $f_{\theta^*}$
    \Statex \hspace{-17pt}\textbf{Client-side representation extraction:}
    \ForAll{clients $i \in \{1,\ldots,N\}$ \textbf{in parallel}}
    \ForAll{samples $x_{i,j} \in \mathcal{D}_i$, $j=1,\ldots,n_i$}
    \State $r_{i,j} \gets E(x_{i,j})/\|E(x_{i,j})\|_2$
    \EndFor
    \State $r_i^* \gets \frac{\sum_j r_{i,j}}{\|\sum_j r_{i,j}\|_2}$, $\rho_i \gets \left\|\frac{1}{n_i}\sum_j r_{i,j}\right\|_2$
    \State Upload $(r_i^*,\rho_i)$ to the server once
    \EndFor
    \Statex \hspace{-17pt}\textbf{Server-side generation and training:}
    \State Initialize the synthetic dataset $\widetilde{\mathcal{D}} \gets \emptyset$
    \ForAll{clients $i \in \{1,\ldots,N\}$}
    \State $\lambda_i \gets \frac{\rho_i^{\gamma}}{\rho_i^{\gamma}+(1-\rho_i)^{\gamma}}$
    \ForAll{classes $c \in \{1,\ldots,C\}$}
    \While{retained samples $< B$}
    \State $\tilde{x} \gets G(t_c,r_i^*,\lambda_i)$
    \State Compute $p(\cdot\mid\tilde{x})$ and $H(\tilde{x})$ using CLIP
    \If{correctly classified and $H(\tilde{x})\leq\tau$}
    \State $\widetilde{\mathcal{D}} \gets \widetilde{\mathcal{D}}\cup\{(\tilde{x},c)\}$
    \EndIf
    \EndWhile
    \EndFor
    \EndFor
    \State Train $f_{\theta}$ on $\widetilde{\mathcal{D}}$ to obtain $f_{\theta^*}$
    \State \Return $f_{\theta^*}$
  \end{algorithmic}
\end{algorithm}


\subsection{Domain Representation Extraction}
\label{sec:domain_representation}

\begin{figure}[t]
  \centering
  \includegraphics[width=0.9\linewidth]{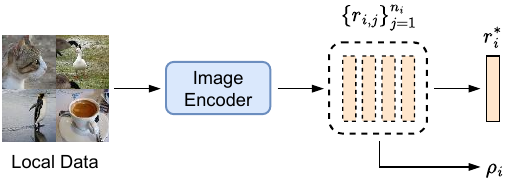}
  \caption{
    Illustration of domain representation extraction for client $i$.
    Local images are encoded into feature vectors $\{ r_{i,j} \}_{j=1}^{n_i}$ by a frozen CLIP image encoder.
    These features are aggregated to obtain a domain representation $r_i^*$.
    The domain concentration score $\rho_i$ is then computed based on the distribution of $\{ r_{i,j} \}_{j=1}^{n_i}$.
  }
  \label{fig:domain_representation}
\end{figure}

To enable lightweight client-side processing while providing explicit visual guidance for domain generation, FedDoG extracts compact domain representations that summarize client-specific visual characteristics.
The overall client-side extraction process is illustrated in \cref{fig:domain_representation}.

Specifically, each client employs a frozen CLIP image encoder to extract semantic features from its local dataset.
For each client $i \in \{1,\ldots,N\}$, we denote its local dataset as $\mathcal{D}_i = \{x_{i,j}\}_{j=1}^{n_i}$, where $x_{i,j}$ denotes the $j$-th sample of client $i$.
For each sample $x_{i,j}$, we compute its normalized feature as
\begin{equation}
  r_{i,j} = \mathrm{CLIP}(x_{i,j}),
\end{equation}
where $\mathrm{CLIP}(\cdot)$ denotes the frozen image encoder with $\ell_2$ normalization, so its output $r_{i,j} \in \mathbb{R}^d$ satisfies $\|r_{i,j}\|_2=1$.
For specialized applications such as medical imaging, domain-specific vision-language encoders have been developed for pathology and broader biomedical data~\cite{huang2023visual,lu2024visual,zhang2025multimodal}.
Using such an encoder in place of a general-purpose CLIP model may yield features better aligned with the target domain and downstream tasks.

Due to CLIP's contrastive pre-training objective, samples with similar semantic characteristics are mapped to nearby locations in the feature space.
Based on this property, we seek a \emph{domain representation} $r_i^*$ that captures the dominant semantic direction of the client's local data by maximizing its average cosine
similarity to local features.
Formally, we define $r_i^*$ as the solution to the following optimization problem:
\begin{equation}
  \label{eq:cosine_objective}
  r^*_i = \arg\max_{\|r\|_2=1} \frac{1}{n_i}\sum_{j=1}^{n_i} \phi(r, r_{i,j}),
\end{equation}
where $\phi(\cdot, \cdot)$ denotes the cosine similarity. The following theorem shows that this problem admits a simple closed-form solution.

\begin{theorem}[Optimal domain representation]
  \leavevmode
  \label{thm:domain_representation}
  For a fixed client $i$, let $\{r_{i,j}\}_{j=1}^{n_i}$ denote the $\ell_2$-normalized feature vectors in $\mathbb{R}^d$.
  If their sum is nonzero, the optimization problem in \cref{eq:cosine_objective} has the following closed-form solution:
  \begin{equation}
    \label{eq:optimal_domain_representation}
    r_i^* = \frac{\sum_{j=1}^{n_i} r_{i,j}}{\left\|\sum_{j=1}^{n_i} r_{i,j}\right\|_2}.
  \end{equation}
  If their sum is zero, every unit vector in $\mathbb{R}^d$ is an optimal solution.
\end{theorem}

The proof of \cref{thm:domain_representation} is provided in Appendix~\ref{app:domain_representation_proof}.
The resulting vector $r_i^*$ in~\cref{eq:optimal_domain_representation} serves as the domain representation of client $i$, which is used to guide server-side conditional data generation.
This representation can be computed efficiently on the client, requiring only one forward pass per sample through a frozen encoder and no iterative optimization.
Because the domain representation is derived solely from image features, the standard FedDoG does not require client labels and is unaffected by missing or noisy annotations.
When class labels are available, the domain granularity can be further refined to the class level by aggregating image features within each class.
We refer to this class-level instantiation as FedDoG-C, in which each client uploads one representation and concentration score per observed class in the same communication round.

Moreover, the geometric distribution of local features reflects the internal consistency of client data, which we further exploit to adaptively control the strength of domain conditioning during generation.

\subsection{Domain-Aware Adaptive Conditioning}
\label{sec:adaptive_conditioning}

Since client domains may exhibit different degrees of concentration in the representation space, a fixed conditioning strength $\lambda$ suboptimal for all clients.
To address this issue, we propose \emph{domain-aware adaptive conditioning}, which adjusts the domain conditioning strength according to the concentration of each client domain.
Intuitively, more concentrated domain features favor stronger visual conditioning to preserve domain-specific characteristics, while less concentrated features benefit from weaker conditioning to promote greater diversity in generated samples.

For each client $i$, we compute the \emph{domain concentration score} $\rho_i$ using the \emph{mean resultant length} (MRL), which is computationally efficient compared with pairwise similarity alternatives.
Formally,
\begin{equation}
  \rho_i = \left\| \frac{1}{n_i}\sum_{j=1}^{n_i} r_{i,j} \right\|_2.
\end{equation}
A larger $\rho_i$ indicates that local features are more concentrated in the representation space and can be better characterized by the domain representation $r_i^*$.

\begin{figure}[t]
  \centering
  \includegraphics[width=0.6\linewidth]{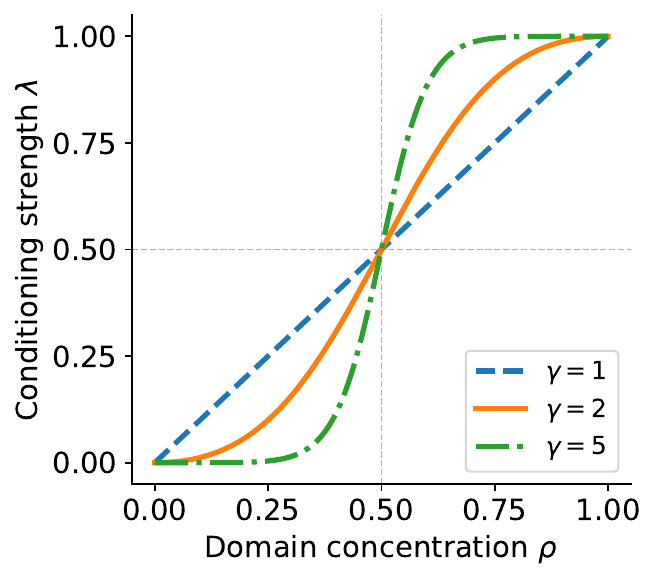}
  \caption{Illustration of the IP-Adapter conditioning strength $\lambda_i$ as a function of the domain concentration score $\rho_i$ for different values of $\gamma$.}
  \label{fig:gating_function}
\end{figure}

We determine the \emph{IP-Adapter conditioning strength} $\lambda_i$ based on the \emph{domain concentration score} $\rho_i$ as follows:
\begin{equation}
  \lambda_i = \frac{\rho_i^{\gamma}}{\rho_i^{\gamma} + (1-\rho_i)^{\gamma}},
  \label{eq:adaptive_strength}
\end{equation}
where $\gamma$ controls the sensitivity of the mapping.
Notably, \cref{eq:adaptive_strength} is equivalent to $\sigma(\gamma \log \frac{\rho_i}{1-\rho_i})$, where $\sigma(\cdot)$ denotes the sigmoid function.
As shown in \cref{fig:gating_function}, a larger $\gamma$ makes the mapping from domain concentration $\rho_i$ to conditioning strength $\lambda_i$ steeper around $\rho_i=0.5$.

\subsection{Domain Generation}
\label{sec:domain_generation}

\begin{figure}[t]
  \centering
  \includegraphics[width=0.9\linewidth]{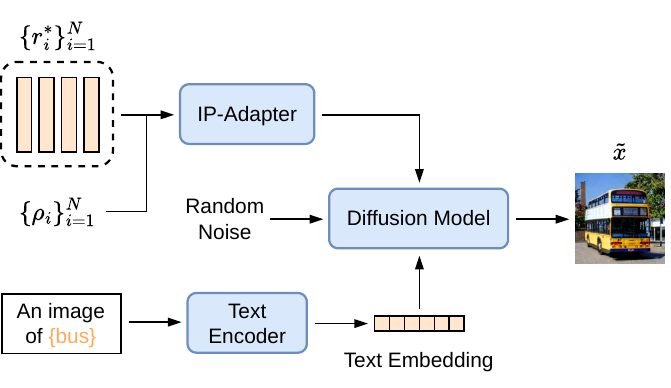}
  \caption{
    Illustration of server-side domain-conditioned image generation.
    Text prompts are encoded separately by a text encoder, while random noise initializes the diffusion process.
  }
  \label{fig:domain_generation}
\end{figure}

To leverage the extracted domain representations $\{r_i^*\}^N_{i=1}$ for global model learning, FedDoG synthesizes images that reflect both target-class semantics and client-specific visual characteristics.
As illustrated in \cref{fig:domain_generation}, this process is performed on the server using a pre-trained diffusion model.

Specifically, we employ a pre-trained latent diffusion model equipped with IP-Adapter, where each client's domain representation $r_i^*$ is incorporated as a visual conditioning signal.
As described in \cref{eq:ip_adapter}, visual conditioning is integrated into the denoising process through decoupled cross-attention.
The corresponding conditioning strength $\lambda_i$ is determined by \cref{eq:adaptive_strength}.
In parallel, class-specific text prompts are encoded by the diffusion model's pre-trained text encoder to provide semantic guidance.

For each client $i$ and target class $c$, the server generates synthetic images under the corresponding domain representation condition and text prompt.
The generated samples are collected to construct the synthetic dataset $\widetilde{\mathcal{D}}$ for global model training.
For FedDoG-C, the server simply replaces the client-level domain representation and concentration score with their class-level counterparts during generation.
Detailed formulations are provided in Appendix~\ref{app:fedogc}.


\subsection{Synthetic Data Filtering}
\label{sec:synthetic_filtering}

Despite domain-conditioned generation, diffusion models may occasionally produce synthetic data with ambiguous or weak semantic content.
To improve the quality and semantic consistency of the generated data, we propose a \emph{synthetic data filtering} strategy on the server.

Specifically, for a generated data sample $\tilde{x}$ with a target class $y$, we compute its similarities to class text prompts $\{t_c\}_{c=1}^C$ using CLIP, and obtain a predictive distribution:
\begin{equation}
  p(c \mid \tilde{x}) = \frac{\exp(\mathrm{CLIP}(\tilde{x}, t_c))}{\sum_{c'} \exp(\mathrm{CLIP}(\tilde{x}, t_{c'}))}.
\end{equation}
Based on this, we perform two filtering steps: correctness filtering and confidence filtering.

\noindent\textbf{Correctness Filtering.}
We retain generated samples whose CLIP-predicted classes match their target labels, i.e.,
$  \arg\max_c p(c \mid \tilde{x}) = y,$
ensuring consistency between the generated data and the target class.

\noindent\textbf{Confidence Filtering.}
Among the correctly classified samples, we further measure semantic confidence using the class-normalized entropy of the predictive distribution:
$  H(\tilde{x}) = - \frac{1}{\log C} \sum_{c=1}^C p(c \mid \tilde{x}) \log p(c \mid \tilde{x}).$
We discard samples whose predictive entropy exceeds the threshold $\tau$, while retaining the remaining samples for server-side global model training. Filtered samples are regenerated until the desired number of valid synthetic data samples is obtained.

%% file: sec/4_experiments.tex
\section{Experiments}
\label{sec:experiments}

\subsection{Setup}
\label{sec:setup}

\textbf{Datasets.}
We evaluate FedDoG under feature distribution skew using three multi-domain benchmarks: PACS~\cite{li2017deeper}, OfficeHome~\cite{venkateswara2017deep}, and DomainNet~\cite{peng2019moment}.
PACS comprises four domains: \emph{art painting} (A), \emph{cartoon} (C), \emph{photo} (P), and \emph{sketch} (S).
OfficeHome comprises four domains: \emph{art} (A), \emph{clipart} (C), \emph{product} (P), and \emph{real world} (R).
DomainNet comprises six domains: \emph{clipart} (C), \emph{infograph} (I), \emph{painting} (P), \emph{quickdraw} (Q), \emph{real} (R), and \emph{sketch} (S).
Following \cite{chen2023fraug, chen2025fedbip}, we use 16, 8, and 16 shots per class for each client on PACS, OfficeHome, and DomainNet, respectively.

\noindent\textbf{Compared Methods.}
We compare FedDoG and FedDoG-C with six baselines covering different OSFL paradigms.
FedD3~\cite{song2023federated} employs dataset distillation to construct compact local datasets, while DENSE~\cite{zhang2022dense} performs data-free knowledge distillation from locally trained models.
FedDEO~\cite{yang2024feddeo}, FedLMG~\cite{yang2025oneshot}, and FedBiP~\cite{chen2025fedbip} leverage pre-trained diffusion models for synthetic data generation.
Additionally, we include a Prompts Only baseline, which generates synthetic images using only class-name prompts without client-specific information.

\noindent\textbf{Implementation Details.}
Each domain serves as one client, yielding $4$ clients for each of PACS and OfficeHome and $6$ clients for DomainNet.
To ensure a fair comparison, we allocate each method other than FedD3 a synthetic-data budget equal to twice the total number of original training samples across all clients.
For FedD3, we use two distilled images per class.
We use Stable Diffusion v1.5~\cite{rombach2022high} as the pretrained diffusion model to generate synthetic images.
For FedDoG, we set $\gamma=2$ on DomainNet, and $\gamma=1$ on PACS and OfficeHome. For FedDoG-C, we set $\gamma=1$ across all three datasets.
We use a confidence threshold of $\tau=0.3$ for filtering.
For all methods, the server trains a ResNet-18 classifier initialized with pretrained weights~\cite{he2016deep} for $50$ epochs using stochastic gradient descent (SGD).
Additional implementation details are provided in Appendix~\ref{app:add_details}.


\subsection{Performance Comparison}
\label{sec:performance_comparison}

\begin{table*}[t]
  \caption{Classification accuracy (\%) under OSFL on PACS, OfficeHome, and DomainNet.
    Results are reported as mean $\pm$ standard deviation for each domain and averaged across domains.
  The best and second-best results are highlighted in \textbf{bold} and \underline{underlined}, respectively.}
  \label{tab:acc_osfl}
  \centering
  \small
  \resizebox{\textwidth}{!}{%
    \begin{tabular}{c|c|c|ccccc|cc}
      \toprule
      Dataset & Domain & Prompts Only & FedD3 & DENSE & FedDEO & FedLMG & FedBiP & FedDoG & FedDoG-C \\
      \midrule
      \multirow{5}{*}{PACS}
      & A   & 51.94\scriptsize{$\pm$1.13} & 58.78\scriptsize{$\pm$3.14} & 65.18\scriptsize{$\pm$3.57} & 53.42\scriptsize{$\pm$3.74} & \underline{65.62}\scriptsize{$\pm$2.71} & 55.80\scriptsize{$\pm$2.48} & 57.74\scriptsize{$\pm$2.11} & \textbf{70.83}\scriptsize{$\pm$2.06} \\
      & C   & 46.13\scriptsize{$\pm$2.73} & 64.59\scriptsize{$\pm$1.80} & 51.04\scriptsize{$\pm$8.19} & 48.21\scriptsize{$\pm$1.95} & 53.87\scriptsize{$\pm$0.93} & \underline{68.90}\scriptsize{$\pm$0.93} & 65.33\scriptsize{$\pm$1.03} & \textbf{75.74}\scriptsize{$\pm$0.68} \\
      & P   & \underline{95.83}\scriptsize{$\pm$0.68} & 90.77\scriptsize{$\pm$3.14} & 84.38\scriptsize{$\pm$2.32} & 94.05\scriptsize{$\pm$1.12} & \textbf{98.07}\scriptsize{$\pm$0.51} & 84.97\scriptsize{$\pm$2.83} & 87.05\scriptsize{$\pm$1.18} & 87.95\scriptsize{$\pm$4.26} \\
      & S   & 31.10\scriptsize{$\pm$4.24} & 60.57\scriptsize{$\pm$2.87} & 33.48\scriptsize{$\pm$5.27} & 32.44\scriptsize{$\pm$2.06} & 40.18\scriptsize{$\pm$4.64} & 35.56\scriptsize{$\pm$7.86} & \underline{64.88}\scriptsize{$\pm$2.54} & \textbf{67.41}\scriptsize{$\pm$3.22} \\
      \cline{2-10}
      & Avg & 56.25\scriptsize{$\pm$1.45} & 68.68\scriptsize{$\pm$0.65} & 58.52\scriptsize{$\pm$1.01} & 57.03\scriptsize{$\pm$0.62} & 64.43\scriptsize{$\pm$1.04} & 61.31\scriptsize{$\pm$0.98} & \underline{68.75}\scriptsize{$\pm$0.51} & \textbf{75.48}\scriptsize{$\pm$2.32} \\
      \midrule
      \multirow{5}{*}{\shortstack{Office\\Home}}
      & A   & 46.62\scriptsize{$\pm$1.24} & 34.37\scriptsize{$\pm$3.48} & \textbf{51.19}\scriptsize{$\pm$0.34} & 46.09\scriptsize{$\pm$2.22} & \underline{51.02}\scriptsize{$\pm$1.31} & 44.67\scriptsize{$\pm$1.31} & 44.40\scriptsize{$\pm$2.68} & 50.09\scriptsize{$\pm$2.36} \\
      & C   & 43.70\scriptsize{$\pm$1.73} & 32.29\scriptsize{$\pm$3.55} & 47.64\scriptsize{$\pm$1.61} & 44.72\scriptsize{$\pm$0.63} & 49.84\scriptsize{$\pm$0.45} & 46.28\scriptsize{$\pm$2.64} & \underline{51.01}\scriptsize{$\pm$0.04} & \textbf{53.40}\scriptsize{$\pm$0.76} \\
      & P   & 72.25\scriptsize{$\pm$1.61} & 48.33\scriptsize{$\pm$2.37} & \underline{75.61}\scriptsize{$\pm$0.72} & 72.76\scriptsize{$\pm$0.47} & 75.33\scriptsize{$\pm$0.62} & 73.57\scriptsize{$\pm$1.60} & 71.91\scriptsize{$\pm$0.40} & \textbf{77.80}\scriptsize{$\pm$1.15} \\
      & R   & 73.58\scriptsize{$\pm$0.74} & 46.88\scriptsize{$\pm$2.86} & \underline{74.80}\scriptsize{$\pm$0.48} & 73.22\scriptsize{$\pm$0.94} & \textbf{77.44}\scriptsize{$\pm$0.85} & 70.63\scriptsize{$\pm$1.57} & 73.99\scriptsize{$\pm$1.15} & 74.44\scriptsize{$\pm$0.65} \\
      \cline{2-10}
      & Avg & 59.04\scriptsize{$\pm$1.27} & 40.47\scriptsize{$\pm$1.18} & 62.31\scriptsize{$\pm$0.20} & 59.20\scriptsize{$\pm$0.71} & \underline{63.41}\scriptsize{$\pm$0.75} & 58.78\scriptsize{$\pm$1.16} & 60.32\scriptsize{$\pm$0.97} & \textbf{63.93}\scriptsize{$\pm$0.73} \\
      \midrule
      \multirow{7}{*}{\shortstack{Domain\\Net}}
      & C   & 56.39\scriptsize{$\pm$0.85} & 52.65\scriptsize{$\pm$3.22} & 68.79\scriptsize{$\pm$1.50} & 57.07\scriptsize{$\pm$0.75} & 64.67\scriptsize{$\pm$1.23} & 65.73\scriptsize{$\pm$1.87} & \underline{76.20}\scriptsize{$\pm$0.84} & \textbf{78.44}\scriptsize{$\pm$0.10} \\
      & I   & 27.81\scriptsize{$\pm$1.29} & 22.29\scriptsize{$\pm$2.61} & 30.52\scriptsize{$\pm$0.73} & 26.60\scriptsize{$\pm$0.65} & 31.95\scriptsize{$\pm$0.67} & 31.06\scriptsize{$\pm$2.29} & \underline{36.54}\scriptsize{$\pm$0.64} & \textbf{36.72}\scriptsize{$\pm$0.44} \\
      & P   & 55.17\scriptsize{$\pm$0.32} & 42.51\scriptsize{$\pm$0.96} & \textbf{60.37}\scriptsize{$\pm$0.49} & 54.09\scriptsize{$\pm$0.12} & \underline{59.48}\scriptsize{$\pm$0.68} & 54.06\scriptsize{$\pm$0.55} & 58.19\scriptsize{$\pm$1.42} & 59.24\scriptsize{$\pm$1.38} \\
      & Q   & 6.48\scriptsize{$\pm$0.94} & 23.89\scriptsize{$\pm$0.82} & 16.12\scriptsize{$\pm$3.63} & 5.37\scriptsize{$\pm$0.25} & 9.29\scriptsize{$\pm$3.86} & 17.77\scriptsize{$\pm$0.81} & \underline{33.40}\scriptsize{$\pm$0.73} & \textbf{35.97}\scriptsize{$\pm$2.18} \\
      & R   & 84.51\scriptsize{$\pm$0.49} & 62.49\scriptsize{$\pm$3.03} & \textbf{87.34}\scriptsize{$\pm$0.61} & 83.87\scriptsize{$\pm$0.90} & \underline{86.72}\scriptsize{$\pm$0.24} & 79.84\scriptsize{$\pm$0.54} & 80.69\scriptsize{$\pm$1.54} & 84.82\scriptsize{$\pm$0.65} \\
      & S   & 49.87\scriptsize{$\pm$0.95} & 41.36\scriptsize{$\pm$1.90} & \textbf{65.05}\scriptsize{$\pm$1.70} & 48.55\scriptsize{$\pm$3.90} & 55.74\scriptsize{$\pm$1.46} & 58.67\scriptsize{$\pm$2.75} & \underline{63.77}\scriptsize{$\pm$1.84} & 63.03\scriptsize{$\pm$1.96} \\
      \cline{2-10}
      & Avg & 46.69\scriptsize{$\pm$0.30} & 40.87\scriptsize{$\pm$0.58} & 54.70\scriptsize{$\pm$1.14} & 45.93\scriptsize{$\pm$0.58} & 51.31\scriptsize{$\pm$0.95} & 51.19\scriptsize{$\pm$1.08} & \underline{58.13}\scriptsize{$\pm$0.82} & \textbf{59.70}\scriptsize{$\pm$0.89} \\
      \bottomrule
    \end{tabular}%
  }
\end{table*}

We evaluate FedDoG under OSFL with feature distribution skew on PACS, OfficeHome, and DomainNet, with results summarized in \cref{tab:acc_osfl}.
Each experiment is repeated three times, and we report the mean and standard deviation across runs.
FedDoG delivers strong performance, ranking second among the compared methods on PACS and DomainNet. It outperforms the competing baselines by up to $3.43$ percentage points. This demonstrates that client-level domain representations provide effective distribution-specific guidance without requiring local class-level statistics.
FedDoG-C achieves the best results on all three benchmarks, improving over FedDoG by $6.73$, $3.61$, and $1.57$ percentage points on PACS, OfficeHome, and DomainNet, respectively. This suggests that class-level representations can provide additional benefits when local labels are available, while FedDoG remains a strong option when only client-level representations are used.


\subsection{Synthetic Data Quality}
\label{sec:synthetic_data_quality}

We evaluate the distributional similarity between synthetic and real data using the CLIP Maximum Mean Discrepancy (CMMD) metric~\cite{jayasumana2024rethinking}.
CMMD computes the maximum mean discrepancy between their CLIP feature distributions, with a lower score indicating that the generated data more closely match the real data distribution.
FedD3 is excluded because its distilled images target downstream utility rather than perceptual fidelity.
Results are reported in \cref{tab:cmmd}.
FedDoG achieves lower average CMMD than all competing baselines across the three datasets, demonstrating the effectiveness of domain representations in capturing domain-specific visual characteristics.
FedDoG-C further improves distributional fidelity and achieves the lowest average CMMD on each dataset, showing the added benefit of incorporating class-level representations.

\begin{table*}[t]
  \caption{CMMD between synthetic and real data on PACS, OfficeHome, and DomainNet.
  Results are reported for each domain and averaged across domains. Lower is better. The best and second-best results are highlighted in \textbf{bold} and \underline{underlined}, respectively.}
  \label{tab:cmmd}
  \centering
  \small
  \begin{tabular}{c|c|c|cccc|cc}
    \toprule
    Dataset & Domain & Prompts Only & DENSE & FedDEO & FedLMG & FedBiP & FedDoG & FedDoG-C \\
    \midrule
    \multirow{5}{*}{PACS}
    & A   & 2.27\scriptsize{$\pm$0.04} & 4.81\scriptsize{$\pm$0.06} & 2.26\scriptsize{$\pm$0.02} & 2.17\scriptsize{$\pm$0.06} & 1.77\scriptsize{$\pm$0.04} & \underline{1.22}\scriptsize{$\pm$0.02} & \textbf{1.11}\scriptsize{$\pm$0.04} \\
    & C   & 2.35\scriptsize{$\pm$0.02} & 5.36\scriptsize{$\pm$0.07} & 2.26\scriptsize{$\pm$0.06} & 2.18\scriptsize{$\pm$0.04} & \underline{1.18}\scriptsize{$\pm$0.19} & 1.20\scriptsize{$\pm$0.01} & \textbf{1.03}\scriptsize{$\pm$0.03} \\
    & P   & 1.96\scriptsize{$\pm$0.02} & 5.21\scriptsize{$\pm$0.07} & 1.98\scriptsize{$\pm$0.03} & 1.86\scriptsize{$\pm$0.01} & 1.82\scriptsize{$\pm$0.15} & \underline{1.48}\scriptsize{$\pm$0.03} & \textbf{1.11}\scriptsize{$\pm$0.04} \\
    & S   & 3.23\scriptsize{$\pm$0.03} & 4.39\scriptsize{$\pm$0.11} & 3.15\scriptsize{$\pm$0.04} & 2.66\scriptsize{$\pm$0.02} & 3.39\scriptsize{$\pm$0.23} & \underline{1.87}\scriptsize{$\pm$0.02} & \textbf{1.84}\scriptsize{$\pm$0.01} \\
    \cline{2-9}
    & Avg & 2.45\scriptsize{$\pm$0.02} & 4.95\scriptsize{$\pm$0.07} & 2.41\scriptsize{$\pm$0.03} & 2.22\scriptsize{$\pm$0.01} & 2.04\scriptsize{$\pm$0.06} & \underline{1.44}\scriptsize{$\pm$0.02} & \textbf{1.27}\scriptsize{$\pm$0.03} \\
    \midrule
    \multirow{5}{*}{\shortstack{Office\\Home}}
    & A   & 2.16\scriptsize{$\pm$0.03} & 4.39\scriptsize{$\pm$0.05} & 2.23\scriptsize{$\pm$0.07} & 1.96\scriptsize{$\pm$0.03} & 1.89\scriptsize{$\pm$0.04} & \underline{1.73}\scriptsize{$\pm$0.01} & \textbf{1.36}\scriptsize{$\pm$0.01} \\
    & C   & 2.54\scriptsize{$\pm$0.04} & 3.90\scriptsize{$\pm$0.06} & 2.53\scriptsize{$\pm$0.03} & 2.28\scriptsize{$\pm$0.03} & 1.83\scriptsize{$\pm$0.09} & \underline{1.27}\scriptsize{$\pm$0.04} & \textbf{1.18}\scriptsize{$\pm$0.01} \\
    & P   & 3.03\scriptsize{$\pm$0.02} & 6.70\scriptsize{$\pm$0.04} & 2.95\scriptsize{$\pm$0.03} & 2.80\scriptsize{$\pm$0.05} & \underline{2.41}\scriptsize{$\pm$0.13} & 2.64\scriptsize{$\pm$0.03} & \textbf{2.17}\scriptsize{$\pm$0.05} \\
    & R   & 1.54\scriptsize{$\pm$0.01} & 4.84\scriptsize{$\pm$0.06} & 1.59\scriptsize{$\pm$0.04} & 1.44\scriptsize{$\pm$0.04} & 1.79\scriptsize{$\pm$0.09} & \underline{1.27}\scriptsize{$\pm$0.03} & \textbf{1.08}\scriptsize{$\pm$0.05} \\
    \cline{2-9}
    & Avg & 2.32\scriptsize{$\pm$0.01} & 4.96\scriptsize{$\pm$0.05} & 2.32\scriptsize{$\pm$0.03} & 2.12\scriptsize{$\pm$0.01} & 1.98\scriptsize{$\pm$0.03} & \underline{1.72}\scriptsize{$\pm$0.01} & \textbf{1.45}\scriptsize{$\pm$0.02} \\
    \midrule
    \multirow{7}{*}{\shortstack{Domain\\Net}}
    & C   & 2.58\scriptsize{$\pm$0.04} & 5.14\scriptsize{$\pm$0.02} & 2.54\scriptsize{$\pm$0.01} & 2.39\scriptsize{$\pm$0.07} & \underline{1.75}\scriptsize{$\pm$0.04} & 1.85\scriptsize{$\pm$0.02} & \textbf{1.30}\scriptsize{$\pm$0.02} \\
    & I   & 3.15\scriptsize{$\pm$0.03} & 6.29\scriptsize{$\pm$0.01} & 3.14\scriptsize{$\pm$0.02} & 2.91\scriptsize{$\pm$0.00} & \underline{2.64}\scriptsize{$\pm$0.10} & 3.02\scriptsize{$\pm$0.03} & \textbf{2.36}\scriptsize{$\pm$0.03} \\
    & P   & 2.50\scriptsize{$\pm$0.05} & 4.84\scriptsize{$\pm$0.01} & 2.50\scriptsize{$\pm$0.03} & 2.22\scriptsize{$\pm$0.01} & \underline{1.37}\scriptsize{$\pm$0.06} & 1.64\scriptsize{$\pm$0.01} & \textbf{1.31}\scriptsize{$\pm$0.03} \\
    & Q   & 5.36\scriptsize{$\pm$0.04} & 4.20\scriptsize{$\pm$0.01} & 5.32\scriptsize{$\pm$0.01} & 4.95\scriptsize{$\pm$0.03} & 3.97\scriptsize{$\pm$0.02} & \textbf{2.57}\scriptsize{$\pm$0.01} & \underline{2.76}\scriptsize{$\pm$0.02} \\
    & R   & 1.59\scriptsize{$\pm$0.02} & 5.88\scriptsize{$\pm$0.02} & 1.58\scriptsize{$\pm$0.02} & 1.52\scriptsize{$\pm$0.01} & \underline{1.42}\scriptsize{$\pm$0.07} & 1.65\scriptsize{$\pm$0.05} & \textbf{1.21}\scriptsize{$\pm$0.01} \\
    & S   & 2.24\scriptsize{$\pm$0.06} & 4.58\scriptsize{$\pm$0.01} & 2.28\scriptsize{$\pm$0.01} & 2.03\scriptsize{$\pm$0.03} & 1.65\scriptsize{$\pm$0.08} & \underline{1.41}\scriptsize{$\pm$0.00} & \textbf{1.24}\scriptsize{$\pm$0.02} \\
    \cline{2-9}
    & Avg & 2.90\scriptsize{$\pm$0.02} & 5.15\scriptsize{$\pm$0.01} & 2.89\scriptsize{$\pm$0.01} & 2.67\scriptsize{$\pm$0.01} & 2.13\scriptsize{$\pm$0.02} & \underline{2.02}\scriptsize{$\pm$0.01} & \textbf{1.70}\scriptsize{$\pm$0.02} \\
    \bottomrule
  \end{tabular}%
\end{table*}

\begin{figure}[t]
  \centering
  \includegraphics[width=\linewidth]{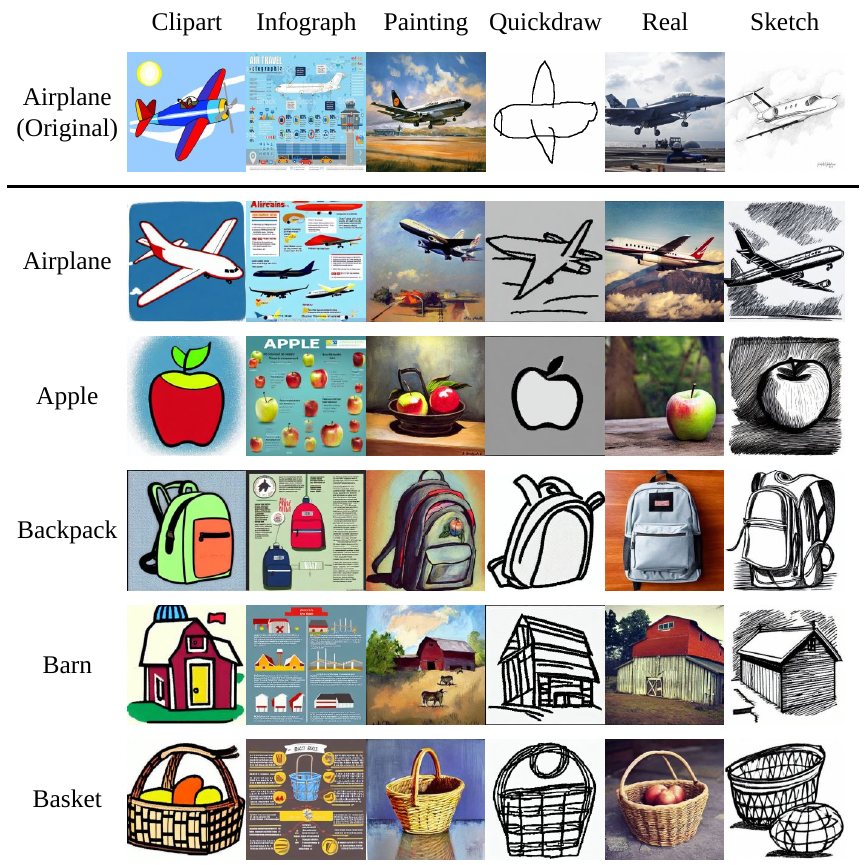}
  \caption{Visualization of original and synthetic images generated by FedDoG on DomainNet.}
  \label{fig:synthetic_images}
\end{figure}

\cref{fig:synthetic_images} compares representative real samples with synthetic images generated by FedDoG on DomainNet.
The synthetic images retain the distinctive visual characteristics of their corresponding domains while remaining semantically consistent with the target classes.
This qualitative evidence complements the quantitative CMMD results, indicating that FedDoG captures domain-specific appearance beyond the semantics provided by class-name prompts.


\subsection{Communication and Computation Cost}
\label{sec:communication_computation}

\begin{table}[t]
  \caption{Communication and computation cost on DomainNet.
  We report the average client upload size, where ``M'' denotes million, and computation time in seconds, denoted by ``s''.}
  \label{tab:communication_cost}
  \centering
  \small
  \begin{tabular}{l|c|c|c}
    \toprule
    & Comm. $\downarrow$ & Comp. $\downarrow$ & Accuracy $\uparrow$ \\
    \midrule
    FedD3 & 7.86M & 27.4s & 40.87 \\
    DENSE & 11.20M & 22.4s & 54.70 \\
    FedDEO & 1.18M & 2572.7s & 45.93 \\
    FedLMG & 11.20M & 26.4s & 51.31 \\
    FedBiP & 5.24M & 317.2s & 51.19 \\
    \midrule
    FedDoG & 0.001M & 1.9s & 58.13 \\
    FedDoG-C & 0.02M & 1.9s & 59.70 \\
    \bottomrule
  \end{tabular}
\end{table}

Because clients in OSFL typically have limited communication bandwidth and computational capacity, we focus on client-side costs, which are key constraints in practical deployment. As shown in \cref{tab:communication_cost}, FedDoG requires an average upload of only $0.001$M, a $99.9\%$ reduction compared with FedDEO ($1.18$M), the most communication-efficient baseline.
FedDoG-C uploads $0.02$M to share class-level representations, still $98.3\%$ less than FedDEO.
Both methods require only $1.9$ seconds of client-side computation, compared with $22.4$ seconds for the fastest baseline, DENSE. This corresponds to a $91.5\%$ reduction in runtime. Their identical reported runtimes suggest that the additional class-wise aggregation in FedDoG-C introduces negligible overhead.
Overall, FedDoG and FedDoG-C substantially reduce client-side communication and computation costs while maintaining strong classification performance.


\subsection{Ablation Study}
\label{sec:ablation}

\begin{table}[t]
  \caption{Test accuracy (\%) of FedDoG under different component combinations.
  ``DAAC'' denotes domain-aware adaptive conditioning.}
  \label{tab:ablation}
  \centering
  \small
  \begin{tabular}{cc|ccc}
    \toprule
    DAAC & Filtering & PACS & OfficeHome & DomainNet \\
    \midrule
    & & 58.71 & 50.94 & 51.09 \\
    $\checkmark$ & & 58.82 & 57.03 & 54.23 \\
    & $\checkmark$ & 63.62 & 60.16 & 55.45 \\
    $\checkmark$ & $\checkmark$ & 68.75 & 60.32 & 58.13 \\
    \bottomrule
  \end{tabular}
\end{table}

We conduct an ablation study on domain-aware adaptive conditioning (DAAC) and synthetic data filtering.
Each component improves test accuracy over the no-component baseline on all three datasets.
Combining both components achieves the highest accuracy on all three datasets, improving over the no-component baseline by up to $10.04$ percentage points.
These results demonstrate that adaptive domain guidance and confidence-based filtering complement each other, improving classifier accuracy consistently across all three benchmarks.


\subsection{Effectiveness of MRL}
\label{sec:mrl_effectiveness}

\begin{figure*}[t]
  \centering
  \includegraphics[width=0.9\linewidth]{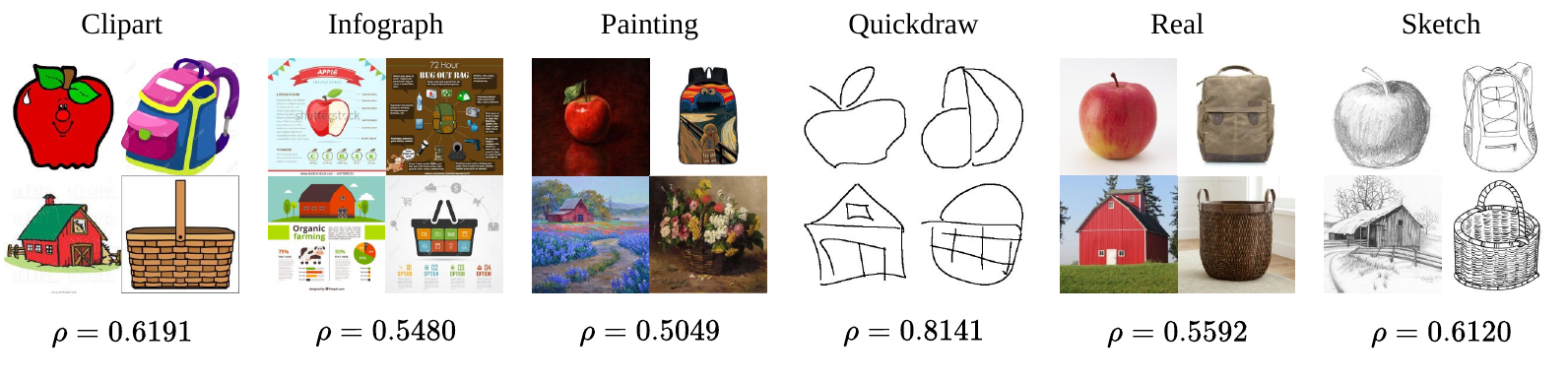}
  \caption{MRL values across different domains on DomainNet.
    For each domain, samples from four classes (apple, backpack, barn, and basket) are shown.
  Some samples are cropped for better visualization.}
  \label{fig:mrl}
\end{figure*}

We analyze the effectiveness of MRL in characterizing the concentration of the client domain.
As shown in \cref{fig:mrl}, different domains exhibit distinct levels of domain concentration.
For example, the \emph{quickdraw} domain achieves a notably higher MRL value, which can be attributed to its highly stylized and abstract visual patterns.
As a result, representations from \emph{quickdraw} tend to cluster tightly in the feature space.
In contrast, domains such as \emph{infograph} and \emph{painting} contain richer textures, diverse color compositions, and substantial stylistic variations, leading to more scattered representations and lower MRL values.
Overall, MRL effectively captures such variations across domains by assigning higher values to more concentrated domains and lower values to more diverse ones.
A comparison with conditioning based on pairwise similarities is provided in Appendix~\ref{app:mrl_pairwise}.


\subsection{Privacy Analysis}
\label{sec:privacy_analysis}

\begin{figure}[t]
  \centering
  \includegraphics[width=\linewidth]{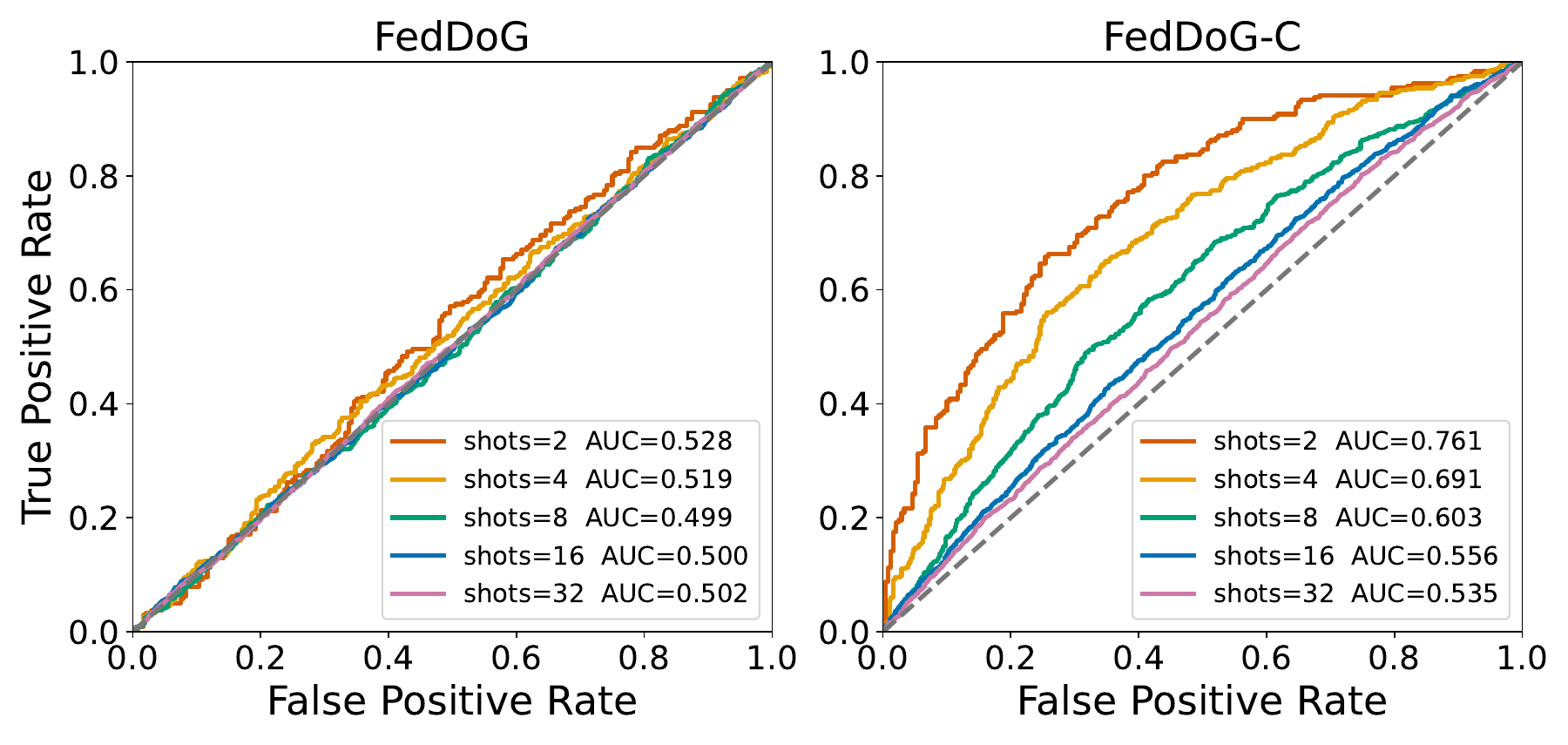}
  \caption{MIA results on DomainNet under different numbers of local shots per class.}
  \label{fig:mia}
\end{figure}

We evaluate membership leakage from the uploaded representations using a LiRA-based attack~\cite{carlini2022membership} adapted to feature aggregation.
The server observes the uploaded representation direction and concentration score and attempts to infer whether a candidate image contributed to a client's upload.
We vary the number of local shots per class from $2$ to $32$ and evaluate balanced sets of member and non-member samples for each client and class.
\cref{fig:mia} reports the resulting ROC curves on DomainNet.
FedDoG's AUC ranges from $0.499$ to $0.528$ across shot counts, indicating that the attack performs close to chance level.
FedDoG-C is more susceptible, particularly at lower shot counts. Its AUC decreases from $0.761$ at $2$ shots to $0.535$ at $32$ shots, suggesting that more local samples reduce the evaluated membership signal.
Overall, increasing the number of local shots reduces the effectiveness of the evaluated MIA, especially for FedDoG-C.
Attack details and additional results are provided in Appendix~\ref{app:privacy}.


\subsection{Sensitivity Analysis of \texorpdfstring{$\gamma$}{gamma}}
\label{sec:gamma_analysis}

We analyze the effect of $\gamma$ on the test accuracy of FedDoG and FedDoG-C on DomainNet.
We evaluate both methods with $\gamma \in \{1,2,3,4,5\}$.
The results are summarized in \cref{tab:gamma_analysis}.
The two variants exhibit different sensitivity to $\gamma$.
FedDoG achieves its highest accuracy of $58.13\%$ at $\gamma=2$, while its accuracy drops at $\gamma=3$ and then partially recovers to $57.87\%$ at $\gamma=5$.
In contrast, FedDoG-C performs best at $\gamma=1$ with an accuracy of $59.70\%$, followed by a gradual decrease as $\gamma$ increases.
FedDoG-C outperforms FedDoG at every tested value of $\gamma$, although the performance gap varies across settings.
These results show that increasing the sensitivity of the conditioning-strength mapping does not consistently improve accuracy, and that the most effective value of $\gamma$ depends on the method variant.

\begin{table}[t]
  \caption{Test accuracy (\%) of FedDoG and FedDoG-C on DomainNet for different values of $\gamma$.}
  \label{tab:gamma_analysis}
  \centering
  \small
  \begin{tabular}{c|ccccc}
    \toprule
    $\gamma$ & 1 & 2 & 3 & 4 & 5 \\
    \midrule
    FedDoG & 57.04 & 58.13 & 55.54 & 57.66 & 57.87 \\
    FedDoG-C & 59.70 & 59.22 & 58.27 & 58.06 & 58.03 \\
    \bottomrule
  \end{tabular}
\end{table}

%% file: sec/5_conclusion.tex
\section{Conclusion}
\label{sec:conclusion}

We have proposed FedDoG, a training-free domain generation framework for OSFL that transfers knowledge from heterogeneous clients through locally extracted domain representations.
Domain-aware adaptive conditioning and synthetic data filtering guide a frozen server-side diffusion model to generate data aligned with heterogeneous client distributions, without optimizing the generator or its conditioning modules.
Experiments on three benchmarks have demonstrated that FedDoG improves synthetic data quality and achieves superior classification accuracy compared with existing OSFL methods, while substantially reducing client-side communication and computation costs.

%% file: sec/6_LLM.tex
\section*{LLM Usage Statement}

ChatGPT (OpenAI) was used to assist with language polishing and improving the clarity and organization of the manuscript throughout the manuscript.
All technical content, experimental results, and scientific conclusions were developed and verified by the authors.

%% file: sec/X_suppl.tex
\clearpage
\setcounter{page}{1}
\maketitlesupplementary

\section{Proof of Theorem~\ref{thm:domain_representation}}
\label{app:domain_representation_proof}

\begin{proof}
  We first note that CLIP features are $\ell_2$-normalized, i.e., $\|r_{i,j}\|_2=1$ for all $j$,
  and the optimization is over unit vectors $\|r\|_2=1$.
  Hence the cosine similarity reduces to an inner product:
  \begin{equation}
    \phi(r, r_{i,j}) = \frac{r^\top r_{i,j}}{\|r\|_2 \, \|r_{i,j}\|_2} = r^\top r_{i,j}.
  \end{equation}
  Let $s_i \triangleq \sum_{j=1}^{n_i} r_{i,j}$ denote the resultant vector. For any feasible $r$, the objective is the following linear functional:
  \begin{align}
    \frac{1}{n_i}\sum_{j=1}^{n_i}\phi(r, r_{i,j})
    &= \frac{1}{n_i}\sum_{j=1}^{n_i} r^\top r_{i,j} \\
    &= \frac{1}{n_i} r^\top s_i.
  \end{align}
  We consider two cases.

  \noindent Case 1: $s_i=0$.
  In this case, the objective equals $r^\top s_i = 0$ for any feasible $r$ with $\|r\|_2=1$.
  Thus any unit vector is optimal.

  \noindent Case 2: $s_i\neq 0$.
  For any unit vector $r$, by the Cauchy--Schwarz inequality,
  \begin{equation}
    r^\top s_i \le \|r\|_2 \|s_i\|_2 = \|s_i\|_2.
    \label{eq:cs_bound}
  \end{equation}
  Thus the objective is upper bounded by $\frac{1}{n_i}\|s_i\|_2$.
  Moreover, the equality condition of Cauchy--Schwarz implies that equality in
  \eqref{eq:cs_bound} holds if and only if $r$ is collinear with $s_i$, and under the unit-norm constraint, the optimal domain representation is given by:
  \begin{equation}
    r_i^*=\frac{s_i}{\|s_i\|_2}
    =\frac{\sum_{j=1}^{n_i} r_{i,j}}{\left\|\sum_{j=1}^{n_i} r_{i,j}\right\|_2}.
  \end{equation}
  This completes the proof.
\end{proof}


\section{FedDoG-C}
\label{app:fedogc}

FedDoG-C is a class-level instantiation of FedDoG when client-side class labels are available.
It follows exactly the same pipeline as FedDoG, including domain representation extraction, domain-aware adaptive conditioning, server-side generation, and synthetic data filtering.
The only difference lies in the granularity at which domain representations and conditioning strengths are defined, which is refined from the client level to the class level.
No additional modules or optimization procedures are introduced.

\subsection{Class-level domain representation}
Let $\mathcal{D}_{i,c}=\{(x,y)\in\mathcal{D}_i \mid y=c\}$ and $n_{i,c}=|\mathcal{D}_{i,c}|$ denote the class-$c$ subset on client $i$.
FedDoG-C computes a class-level domain representation by aggregating CLIP image features within each class:
\begin{equation}
  r_{i,c}^\ast
  = \frac{\sum_{(x,y)\in\mathcal{D}_{i,c}} \mathrm{CLIP}(x)}
  {\left\|\sum_{(x,y)\in\mathcal{D}_{i,c}} \mathrm{CLIP}(x)\right\|_2}.
\end{equation}
This is a direct application of \cref{eq:optimal_domain_representation} to samples belonging to class $c$.
Accordingly, client $i$ uploads $\{r_{i,c}^\ast\}_{c\in\mathcal{C}_i}$ instead of a single $r_i^\ast$.

\subsection{Class-level adaptive conditioning}
FedDoG computes a domain concentration score using MRL and maps it to the IP-Adapter conditioning strength.
FedDoG-C follows the same procedure at the class level:
\begin{equation}
  \rho_{i,c}
  = \left\| \frac{1}{n_{i,c}} \sum_{(x,y)\in\mathcal{D}_{i,c}} \mathrm{CLIP}(x) \right\|,
\end{equation}
which is then mapped to the class-level conditioning strength via the same function:
\begin{equation}
  \lambda_{i,c}
  = \frac{\rho_{i,c}^{\gamma}}{\rho_{i,c}^{\gamma} + (1-\rho_{i,c})^{\gamma}}.
\end{equation}

\subsection{Server-side generation}
FedDoG-C follows the same server-side diffusion model and IP-Adapter mechanism as FedDoG, with class-level conditioning signals replacing client-level ones.
For generating a class-$c$ synthetic sample associated with client $i$, the server conditions the diffusion model using the corresponding class-level domain representation and conditioning strength $(r_{i,c}^\ast, \lambda_{i,c})$.

The server independently uses each client-specific class-level conditioning pair $(r_{i,c}^\ast, \lambda_{i,c})$ to generate class-$c$ synthetic samples, preserving the corresponding client-specific domain characteristics. No aggregation is performed across clients.
All remaining steps, including synthetic data filtering, are identical to those in FedDoG.


\section{Additional Implementation Details}
\label{app:add_details}

\subsection{Experimental Environment}
\label{app:experimental_environment}

All experiments were conducted on a machine running Ubuntu 20.04.6 LTS.
The machine is equipped with an Intel Core i9-13900KF CPU (32 cores), 64 GB of system memory, and an NVIDIA GeForce RTX 4090 GPU with 24 GB of VRAM.
All algorithms are implemented in Python 3.11 using PyTorch 2.7.1 with CUDA 12.8.

\subsection{Implementation Details for OSFL}
\label{app:osfl_details}

\paragraph{Dataset construction.}
For DomainNet, we use the first 20 classes in the official label order across its six domains.
For PACS, we use seven classes: \emph{dog}, \emph{elephant}, \emph{giraffe}, \emph{guitar}, \emph{horse}, \emph{house}, and \emph{person}.
For OfficeHome, we follow the 20-class subset used by FedBiP~\cite{chen2025fedbip}: \emph{Bucket}, \emph{Chair}, \emph{Couch}, \emph{Fan}, \emph{Fork}, \emph{Hammer}, \emph{Kettle}, \emph{Knives}, \emph{Marker}, \emph{Monitor}, \emph{Mop}, \emph{Notebook}, \emph{Oven}, \emph{Pen}, \emph{Pencil}, \emph{Sink}, \emph{Speaker}, \emph{Spoon}, \emph{Telephone}, and \emph{Toys}.
Each domain is assigned to a separate client.
This construction results in six clients with identical label spaces but substantially different feature distributions.

\subsection{Implementation Details for Compared Methods}
\label{app:baseline_details}

\paragraph{Shared server-side classifier.}
For all methods, the server-side classifier is trained with a batch size of 64, a learning rate of 0.01, momentum of 0.9, and weight decay of $10^{-5}$.

\paragraph{FedD3.}
Each client distills 2 images per class using Adam with a learning rate of 0.004 for at most 3,000 optimization steps. At each step, one real image per class is sampled. Optimization stops early when the accuracy reaches 0.999. We use a four-layer fully connected ReLU network to compute the NTK, with a kernel regularization coefficient of $10^{-6}$.

\paragraph{DENSE.}
Clients train for 20 epochs with a batch size of 32 using SGD with a learning rate of 0.01. On the server, synthesis and distillation proceed for 50 rounds, with 30 generator optimization steps per round. The generator uses Adam with a learning rate of 0.001 and a noise dimension of 256. The batch-normalization matching loss, class cross-entropy loss, and adversarial loss each have a weight of 1; the distillation temperature is 20.

\paragraph{FedDEO.}
Client description embeddings are optimized for 200 epochs with a batch size of 4 using SGD with a learning rate of 10 and no weight decay. The class-text and client-description guidance strengths are both set to 3.

\paragraph{FedLMG.}
Each client trains a local ResNet-18 for 20 epochs with a batch size of 32 using SGD with a learning rate of 0.01. The classification loss and batch-normalization statistics loss are weighted by 1 and 0.1, respectively. The latent-guidance step size is 1, the warm-up optimization lasts 10 steps.

\paragraph{FedBiP.}
Client concept vectors are optimized for 50 epochs with a batch size of 8 using Adam with a learning rate of 0.01. The standard deviations of the same-class latent interpolation coefficient and domain-concept perturbation are both set to 0.1. The text-guidance strength is 7.5.

\subsection{Implementation Details for FedDoG}
\label{app:feddog_details}

\paragraph{Diffusion model and IP-Adapter.}
We adopt Stable Diffusion v1.5\footnote{\url{https://huggingface.co/stable-diffusion-v1-5/stable-diffusion-v1-5}} as the server-side diffusion model in all experiments.
To enable image-conditioned generation, we use the \texttt{ip-adapter\_sd15}\footnote{\url{https://huggingface.co/h94/IP-Adapter}}, which incorporates image features into the diffusion process via decoupled cross-attention.
The image encoder used by IP-Adapter is OpenCLIP ViT-H/14.
The same OpenCLIP ViT-H/14 image encoder is also used for domain representation extraction and synthetic data filtering.
Both the diffusion model and the IP-Adapter are kept frozen.
For text conditioning, we use the prompt template ``An image of \{\texttt{class}\}'' for each target class.

\paragraph{Synthetic data filtering.}
To keep server-side generation time bounded and prevent the filtering loop from running indefinitely, we impose a per-position limit on candidate generation.
For each synthetic-data position to be retained, the server evaluates at most 50 candidate images individually.
Consistent with the correctness criterion in \cref{sec:synthetic_filtering}, our experiments require the target class to be the top-1 CLIP prediction and use a class-normalized entropy threshold of $\tau=0.3$.
For tasks with many classes or low inter-class separability, we recommend relaxing the correctness requirement to a Top-$k$ criterion, such as Top-5, under which a candidate is considered correct if its target class appears among the $k$ highest-ranked predictions.
In general, a candidate passes filtering when $\operatorname{rank}_{\tilde{x}}(y)\leq k$ and $H_{\mathrm{norm}}(\tilde{x})\leq\tau$, with $k=1$ in our experiments.
As soon as an accepted candidate is found, it is saved immediately and no further candidates are considered for that position.
If no candidate passes within the 50-image limit, the server saves the best candidate according to the following lexicographic criterion:
\begin{equation}
  \arg\max_{\tilde{x}}
  \left(
    \mathbf{1}[\operatorname{rank}_{\tilde{x}}(y)\leq k],
    p(y\mid\tilde{x}),
    -H_{\mathrm{norm}}(\tilde{x})
  \right),
\end{equation}
where $\operatorname{rank}_{\tilde{x}}(y)$ is the rank of the target class under the CLIP predictions, and ties are broken by higher target-class probability and then lower normalized entropy.
After saving either an accepted candidate or the best fallback, the server clears the candidate records and resets the counter before processing the next position.


\section{Comparison with Pairwise Similarity Conditioning}
\label{app:mrl_pairwise}

We investigate pairwise similarity conditioning as an alternative to the proposed MRL-based approach, using mean pairwise cosine similarity as a representative example.
All experimental settings except the conditioning-strength rule are kept fixed.
Let $(S_i)_{uw}=r_{i,u}^{\top}r_{i,w}$ denote the cosine similarity between normalized features at client $i$.
Excluding self-similarities, the baseline computes
\begin{equation}
  c_i
  = \frac{\sum_{u=1}^{n_i}\sum_{w=1}^{n_i}(S_i)_{uw}
  -\operatorname{tr}(S_i)}{n_i(n_i-1)}.
  \label{eq:pairwise_cosine_mean}
\end{equation}
The mean is mapped to $[0,1]$ and used directly as the IP-Adapter strength, without an additional $\gamma$ mapping:
\begin{equation}
  \lambda_i^{\mathrm{cos}} = q_i
  = \operatorname{clip}\!\left(\frac{c_i+1}{2},\,0,\,1\right).
  \label{eq:pairwise_strength}
\end{equation}
FedDoG retains the mapping in \cref{eq:adaptive_strength} with $\gamma=2$.
\cref{tab:mrl_pairwise_scores} compares the domain concentration scores and corresponding IP-Adapter conditioning strengths across domains.

\begin{table}[t]
  \centering
  \small
  \caption{Domain concentration scores and IP-Adapter conditioning strengths on DomainNet.}
  \label{tab:mrl_pairwise_scores}
  \begin{tabular}{l|cc|cc}
    \toprule
    & \multicolumn{2}{c|}{FedDoG (MRL)} & \multicolumn{2}{c}{Pairwise cosine} \\
    \cmidrule(lr){2-3}\cmidrule(lr){4-5}
    Domain & $\rho_i$ & $\lambda_i^{\mathrm{MRL}}$ & $c_i$ & $q_i=\lambda_i^{\mathrm{cos}}$ \\
    \midrule
    Clipart   & 0.6191 & 0.7254 & 0.3813 & 0.6907 \\
    Infograph & 0.5480 & 0.5950 & 0.2980 & 0.6490 \\
    Painting  & 0.5049 & 0.5098 & 0.2526 & 0.6263 \\
    Quickdraw & 0.8141 & 0.9504 & 0.6617 & 0.8309 \\
    Real      & 0.5592 & 0.6167 & 0.3105 & 0.6553 \\
    Sketch    & 0.6120 & 0.7133 & 0.3726 & 0.6863 \\
    \bottomrule
  \end{tabular}
\end{table}

For normalized features and $n_i>1$, mean pairwise cosine similarity satisfies $c_i=(n_i\rho_i^2-1)/(n_i-1)$.
It can therefore be computed from MRL without constructing the pairwise similarity matrix.
However, this simplification does not generally extend to other nonlinear similarity functions, such as Gaussian RBF similarity, Laplacian similarity based on Euclidean distance, and sigmoid similarity.
Direct evaluation of their exact averages over all sample pairs requires $O(n_i^2d)$ operations.
In contrast, MRL requires only $O(n_id)$ operations through feature summation.
Thus, MRL avoids the quadratic cost of explicitly evaluating these nonlinear similarities over all sample pairs.

\cref{tab:mrl_pairwise_efficiency} reports downstream classification accuracy under the two conditioning rules.
The MRL-based rule achieves a $0.38$ percentage point accuracy improvement over the tested pairwise cosine rule.
The tested pairwise conditioning rule therefore provides no observed accuracy advantage.

\begin{table}[t]
  \centering
  \small
  \caption{Downstream test accuracy (\%) on DomainNet under different conditioning rules. Results are reported as mean $\pm$ standard deviation.}
  \label{tab:mrl_pairwise_efficiency}
  \begin{tabular}{l|c}
    \toprule
    Conditioning rule & Accuracy \\
    \midrule
    FedDoG (MRL)    & 58.13\scriptsize{$\pm$0.82} \\
    Pairwise cosine & 57.75\scriptsize{$\pm$1.18} \\
    \bottomrule
  \end{tabular}
\end{table}


\section{Membership Inference Attack on Uploaded Representations}
\label{app:privacy}

\subsection{Threat Model and Relation to LiRA}
We consider a server that follows the upload protocol but attempts to infer
whether a candidate image $x$ belongs to a particular client's local dataset.
The attacker observes the uploaded representation direction and concentration
score, knows the frozen image encoder and its preprocessing and the class names.
For FedDoG-C, it additionally knows the candidate's
class and attacks the corresponding class-level upload. The FedDoG attack
does not use the candidate's class. Neither the client's total sample count
nor its per-class sample counts are provided to the attacker. Membership
labels are used only for evaluation after attack scores have been computed.

Our attack adapts LiRA's candidate-specific IN/OUT Gaussian modeling and
likelihood-ratio scoring~\cite{carlini2022membership}. Original LiRA uses
shadow classifiers and transformed prediction confidences. Here, we simulate
feature aggregates and model a scalar statistic of the uploaded mean.
We also use references generated from the target upload and marginalize over
unknown aggregation sizes. We therefore refer to the procedure as a
\emph{LiRA-based shadow-aggregation attack}.

\subsection{Attack Construction}
Let $z(x)$ denote the $\ell_2$-normalized feature of $x$ extracted by the
frozen encoder. The direction $r_i^*$ and concentration $\rho_i$ jointly
reveal the empirical mean feature:
\begin{equation}
  \bar z_i = \rho_i r_i^*
  = \frac{1}{n_i}\sum_{j=1}^{n_i} z(x_{i,j}).
  \label{eq:privacy_mean}
\end{equation}
This reconstruction does not require knowing $n_i$.
For FedDoG-C, the same identity holds for each class-level pair
$(r_{i,c}^*,\rho_{i,c})$ and its class-specific mean $\bar z_{i,c}$.

Because the server has neither access to clients' raw data nor knowledge of their domains, it cannot directly sample from each client's true data distribution. Instead, FedDoG uses each client's uploaded representation to generate client-conditioned synthetic images that approximate the distribution and serve as attack references. The generated references are not filtered, and their class labels are assigned according to the generation prompts.
For FedDoG, let $Q_i$ be the empirical distribution of normalized features over the client's generated references, with uniform class weights.
For FedDoG-C, let $Q_{i,c}$ contain only references generated for the candidate's class using its class-level upload.
Below, $Q$ denotes the applicable reference distribution, $\mu=\mathbb{E}_{v\sim Q}[v]$ its empirical center, and $\bar z$ the mean recovered from the attacked upload.

Given a candidate feature $z=z(x)$, we compute
\begin{equation}
  t_x(\bar z)=(z-\mu)^\top(\bar z-\mu).
  \label{eq:privacy_statistic}
\end{equation}
This statistic measures alignment between the centered candidate and
centered uploaded mean. It replaces the classifier-confidence statistic
used in original LiRA.

To estimate IN and OUT score distributions, for each assumed aggregation size $n$, we draw $n-1$ background reference
features $v_1,\ldots,v_{n-1}$ and an independent replacement $v'$ from $Q$,
with replacement. A paired shadow draw is
\begin{equation}
  \begin{aligned}
    a^{\mathrm{IN}}_{n,b}
    &= \frac{z+\sum_{j=1}^{n-1}v_j}{n},\\
    a^{\mathrm{OUT}}_{n,b}
    &= \frac{v'+\sum_{j=1}^{n-1}v_j}{n}.
  \end{aligned}
  \label{eq:privacy_shadows}
\end{equation}
Both aggregates contain $n$ feature vectors, and only one slot changes between the candidate and a reference replacement.
We generate $B=128$ pairs for each size and fit separate scalar Gaussians to $t_x(a^{\mathrm{IN}}_{n,b})$ and $t_x(a^{\mathrm{OUT}}_{n,b})$.
Their empirical means and unbiased sample variances are denoted by $m_{x,n}^{h}$ and $(\sigma_{x,n}^{h})^2$ for $h\in\{\mathrm{IN},\mathrm{OUT}\}$.
Variances are floored at $10^{-8}$.
Bootstrapping increases the number of shadow draws, not the number of independent reference images.

Let $\mathcal{K}=\{2,4,8,16,32\}$ and let $C$ be the known number of
classes in the evaluation subset. We assign a uniform prior over
\begin{equation}
  \mathcal{N}=
  \begin{cases}
    \{Ck:k\in\mathcal{K}\}, & \text{FedDoG},\\
    \mathcal{K}, & \text{FedDoG-C}.
  \end{cases}
  \label{eq:privacy_sizes}
\end{equation}
The prior is fixed across local shot counts and does not read the
target client's actual counts.
The candidate's attack score is the log ratio of mixture densities:
\begin{equation}
  \begin{aligned}
    \ell_x &= \log p_{\mathrm{IN}}(t_x(\bar z))
    -\log p_{\mathrm{OUT}}(t_x(\bar z)),\\
    p_h(t) &= \frac{1}{|\mathcal{N}|}\sum_{n\in\mathcal{N}}
    \mathcal{N}\!\left(t;m_{x,n}^{h},(\sigma_{x,n}^{h})^2\right).
  \end{aligned}
  \label{eq:privacy_likelihood}
\end{equation}
Larger scores favor membership.

\subsection{Evaluation Protocol}

We vary the number of local shots per class, $s\in\{2,4,8,16,32\}$.
Within each client and class, both candidate lists are sorted by path and truncated to the smaller count, yielding equal numbers of members and non-members.
\cref{tab:privacy_counts} gives the resulting counts.
In particular, OfficeHome has eight held-out images
per domain and class, so its candidate count saturates at $s=8$ even when
the local shot count increases.

\begin{table}[t]
  \centering
  \caption{Number of members evaluated at each per-class shot count.
    The number of non-members is identical. Counts are the same for FedDoG
  and FedDoG-C.}
  \small
  \label{tab:privacy_counts}
  \begin{tabular}{lrrrrr}
    \hline
    Dataset & 2 & 4 & 8 & 16 & 32 \\
    \hline
    DomainNet & 120 & 240 & 480 & 956 & 1796 \\
    PACS & 56 & 112 & 224 & 448 & 896 \\
    OfficeHome & 160 & 320 & 640 & 640 & 640 \\
    \hline
  \end{tabular}
\end{table}

\subsection{Additional Results}

\begin{figure}[t]
  \centering
  \includegraphics[width=\linewidth]{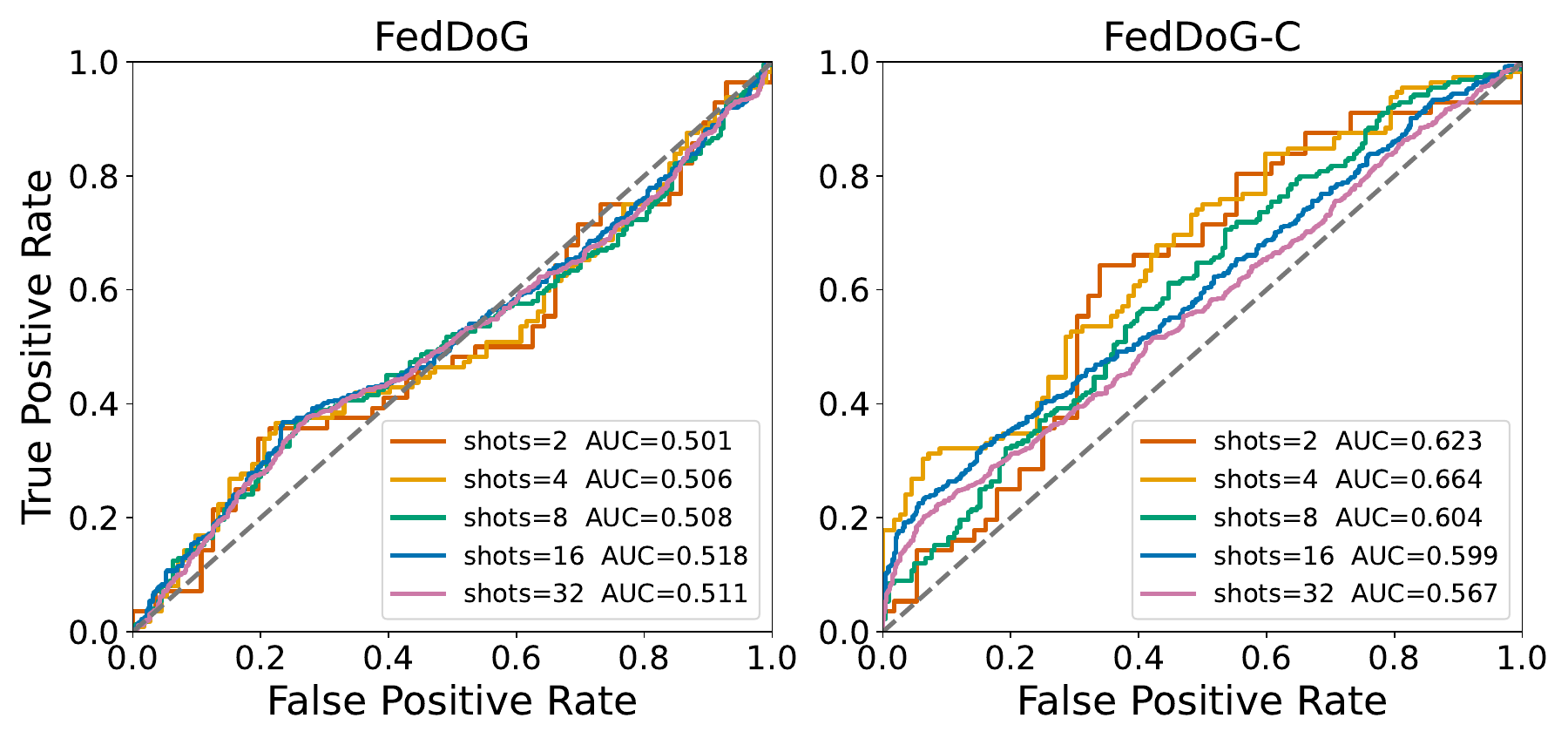}
  \caption{MIA results on PACS under different numbers of local shots per class.}
  \label{fig:mia_pacs}
\end{figure}

\begin{figure}[t]
  \centering
  \includegraphics[width=\linewidth]{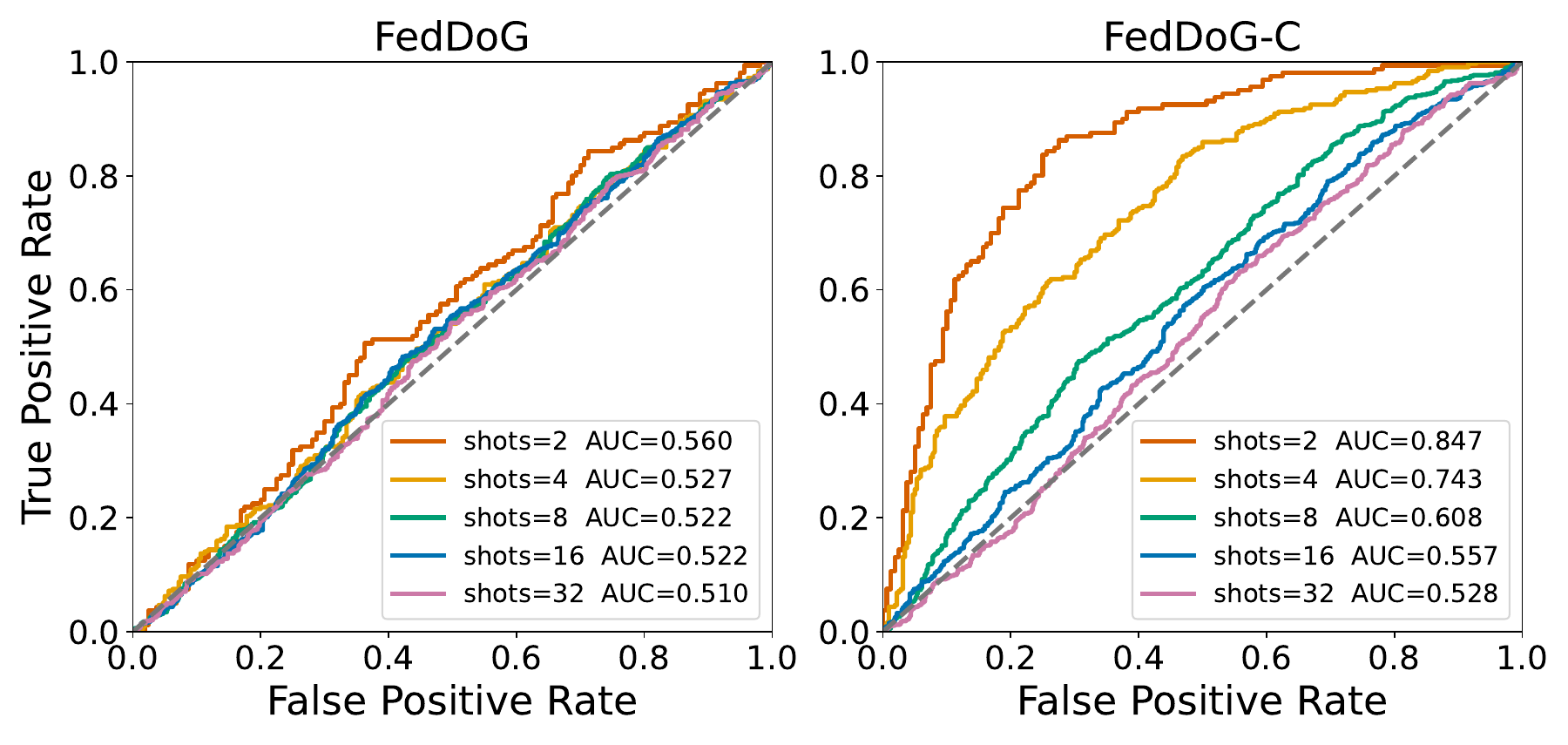}
  \caption{MIA results on OfficeHome under different numbers of local shots per class.}
  \label{fig:mia_officehome}
\end{figure}

\cref{fig:mia_pacs} and~\cref{fig:mia_officehome} report
the additional ROC curves, and \cref{tab:privacy_auc} summarizes AUCs.
On PACS, FedDoG remains close to chance-level overall discrimination,
with AUCs ranging from 0.501 to 0.518. FedDoG-C has higher AUC at every
tested shot count. It peaks at 0.664 at 4 shots before declining to 0.567
at 32 shots, so its AUC varies non-monotonically with shot count.
On OfficeHome, FedDoG's AUC ranges from 0.510 to 0.560, while FedDoG-C's
AUC declines from 0.847 at 2 shots to 0.528 at 32 shots.
Across these datasets, class-level uploads expose more membership signal
to the attack than client-level uploads, particularly at small shot
counts. Increasing the shot count generally lowers attack AUC, although
trends vary by dataset and synthetic-reference calibration.

\begin{table*}[t]
  \centering
  \caption{Micro ROC-AUC of the LiRA-based shadow-aggregation attack.
    Each column gives the local per-class shot count. Larger AUC indicates
  greater overall membership discrimination by the evaluated attack.}
  \small
  \label{tab:privacy_auc}
  \begin{tabular}{l|lccccc}
    \hline
    Dataset & Method & 2 & 4 & 8 & 16 & 32 \\
    \hline
    \multirow{2}{*}{DomainNet} & FedDoG & 0.528 & 0.519 & 0.499 & 0.500 & 0.502 \\
    & FedDoG-C & 0.761 & 0.691 & 0.603 & 0.556 & 0.535 \\
    \midrule
    \multirow{2}{*}{PACS} & FedDoG & 0.501 & 0.506 & 0.508 & 0.518 & 0.511 \\
    & FedDoG-C & 0.623 & 0.664 & 0.604 & 0.599 & 0.567 \\
    \midrule
    \multirow{2}{*}{OfficeHome} & FedDoG & 0.560 & 0.527 & 0.522 & 0.522 & 0.510 \\
    & FedDoG-C & 0.847 & 0.743 & 0.608 & 0.557 & 0.528 \\
    \hline
  \end{tabular}
\end{table*}


\section{Sensitivity Analysis of \texorpdfstring{$\tau$}{tau}}
\label{app:tau_analysis}

\begin{table*}[t]
  \centering
  \small
  \caption{Generation and filtering statistics of FedDoG on DomainNet under different entropy thresholds.}
  \label{tab:tau_analysis}
  \begin{tabular}{l|cccccc}
    \toprule
    Metric & $\tau=0.1$ & $\tau=0.2$ & $\tau=0.3$ & $\tau=0.4$ & $\tau=0.5$ & $\tau=0.6$ \\
    \midrule
    Required samples & 3840 & 3840 & 3840 & 3840 & 3840 & 3840 \\
    Regeneration attempts & 8900 & 6017 & 4639 & 3984 & 3627 & 3466 \\
    Samples reaching regeneration limit & 48 & 24 & 10 & 6 & 5 & 4 \\
    Test accuracy (\%) & 56.63 & 58.35 & 58.13 & 57.19 & 56.92 & 56.44 \\
    \bottomrule
  \end{tabular}
\end{table*}

To study how the entropy threshold affects filtering efficiency, we evaluate FedDoG on DomainNet with $\tau \in \{0.1,0.2,0.3,0.4,0.5,0.6\}$, keeping all other settings fixed, including $k=1$ and the $50$-candidate limit.
For each threshold, we use a fixed target of $3{,}840$ synthetic samples and record the total number of regeneration attempts (candidate images generated after the initial attempt) and the number of samples reaching the regeneration limit and using fallback selection.
\cref{tab:tau_analysis} summarizes these statistics for each threshold.

As $\tau$ increases from $0.1$ to $0.6$, regeneration attempts decrease from $8{,}900$ to $3{,}466$, while the number of samples reaching the regeneration limit decreases from $48$ to $4$, indicating improved generation efficiency at higher thresholds.
Test accuracy varies non-monotonically with $\tau$. It is $56.63\%$ at $\tau=0.1$, peaks at $58.35\%$ at $\tau=0.2$, remains close at $58.13\%$ at $\tau=0.3$, and falls to $56.44\%$ at $\tau=0.6$.
Thus, stricter filtering with a lower threshold does not necessarily improve task performance.
Overall, $\tau=0.3$ offers a favorable trade-off compared with $\tau=0.2$, achieving similar accuracy with fewer regeneration attempts ($4{,}639$ vs. $6{,}017$) and fewer samples reaching the regeneration limit ($10$ vs. $24$).
We therefore use $\tau=0.3$ as the default setting in all experiments.


\section{Effect of the Pre-trained Diffusion Model}
\label{app:diffusion_model}

We evaluate the effect of the server-side diffusion model on DomainNet by comparing Stable Diffusion v1.5 (SD 1.5)~\cite{rombach2022high} with Stable Diffusion XL (SDXL)~\cite{podell2024sdxl} while holding all other experimental settings fixed.
Compared with SD 1.5, SDXL uses a U-Net backbone approximately three times larger than those in earlier Stable Diffusion versions and adds a second text encoder.
It also introduces additional conditioning schemes, is trained across multiple aspect ratios, and supports an optional refinement model that applies post-hoc image-to-image processing to improve sample fidelity.
\cref{tab:diffusion_model_comparison} reports test accuracy on DomainNet for both FedDoG variants.
SDXL improves test accuracy over SD 1.5 by $3.71$ percentage points for FedDoG and $2.75$ percentage points for FedDoG-C.
These gains for both client-level and class-level variants show that the FedDoG pipeline can leverage SDXL without modification and achieve higher accuracy on DomainNet, highlighting the flexibility of FedDoG with respect to the diffusion backbone.

\begin{table}[t]
  \centering
  \small
  \caption{Test accuracy (\%) on DomainNet for FedDoG and FedDoG-C using Stable Diffusion v1.5 and SDXL.}
  \label{tab:diffusion_model_comparison}
  \begin{tabular}{cc|cc}
    \toprule
    \multicolumn{2}{c|}{FedDoG} & \multicolumn{2}{c}{FedDoG-C} \\
    \cmidrule(lr){1-2}\cmidrule(lr){3-4}
    SD 1.5 & SDXL & SD 1.5 & SDXL \\
    \midrule
    58.13 & 61.84 & 59.70 & 62.45 \\
    \bottomrule
  \end{tabular}
\end{table}


\section{Limitations}
\label{app:limitations}

FedDoG summarizes client data distributions using compact semantic representations. This design enables efficient communication and training-free server-side generation, but it may not fully capture highly multi-modal or fine-grained variations within a client. FedDoG-C partially addresses this issue by refining the representation granularity to the class level when labels are available. Exploring more flexible domain granularity or structured representations is a promising future direction.

In addition, FedDoG relies on a fixed pre-trained diffusion model for server-side generation. Although this keeps the framework training-free and efficient, the generated data quality may still depend on the coverage of the pre-trained generative model. Extending FedDoG to more specialized domains or broader modalities remains an interesting direction for future work.